\documentclass{article}
\usepackage{geometry}
\usepackage{booktabs}
\usepackage{graphicx} 
\usepackage{multirow}
\usepackage{multicol}
\usepackage{caption,subcaption}%
\usepackage{calc}
\usepackage{tikz}
\usepackage{setspace}

\tikzset{
    module/.style={%
        draw, rounded corners,
        thick,
        minimum width=20mm,
        align=center,
        minimum height=10mm,
        },
}

\title{Detecting Alarming Student Verbal Responses using Text and Audio Classifier}
\author{Christopher Ormerod and Gitit Kehat}
\date{Paper to be Presented at the National Council on Measurement in Education Conference on April 10, 2026}

\begin{document}

\maketitle

\begin{abstract}
This paper addresses a critical safety gap in the use Automated Verbal Response Scoring (AVRS). We present a novel hybrid framework for troubled student detection that combines a text classifier, trained to detect responses based on their content, and an audio classifier, trained to detect responses using prosodic markers. This approach overcomes key limitations of traditional AVRS systems by considering both content and prosody of responses, achieving enhanced performance in identifying potentially concerning responses. This system can expedite the review process by humans, which can be life-saving particularly when timely intervention may be crucial.
\end{abstract}

\section{Introduction}

Automated Scoring (AS) systems are sophisticated statistical models designed to evaluate student responses, mimicking the grading process of human educators. When AS are held to high standards, these systems have proven to be cost-effective alternatives to hand scoring, making them an increasingly attractive option for educational institutions and assessment organizations dealing with large volumes of student work \cite{williamson_framework_2012}. Examples include Automated Essay Scoring (AES) \cite{shermis_contrasting_2013}, Automated Short Answer Scoring (ASAS) \cite{shermis_contrasting_2015}, and Automated Verbal Response Scoring (AVRS) \cite{cahill_natural_2020}. The benefits of these systems are accompanied by many risks. AS systems can be vulnerable to gaming \cite{lottridge_comparing_nodate}, they can be less accurate on certain classes of responses, and they can perpetuate and introduce additional biases beyond the training data \cite{ormerod_automated_2021}. These biases can be exacerbated for AVRS systems where ASR can be unreliable for certain subgroups \cite{kwako_using_2022}. 

A critical yet often overlooked risk in taking humans out of the hand-scoring process is that humans naturally respond with concern when confronted with a student response that indicates that the student is at risk of self-harm or inflicting harm on others. These responses, in some hand-scoring materials, are simply called "Alerts", which will be the term we will use in this article. In traditional human scoring processes, it might take weeks before such Alerts are noticed, delaying crucial interventions. Recent studies have demonstrated that Large Language Models (LLMs) can effectively flag a small percentage of text responses for immediate human review, significantly expediting the process and enabling timely action when necessary \cite{ormerod_using_2023}. We seek to implement a similar pipeline for verbal student responses.

Traditional AVRS systems typically integrate automatic speech recognition (ASR) with ASAS. However, applying a standard AVRS approach to detect alerts presents two key limitations: ASR systems often struggle with distressed speech and miss crucial vocal indicators. Conversely,  focusing solely on tone can overlook concerning content delivered in a neutral voice. Our work shows that a hybrid detection framework achieves enhanced performance. By a audio classifier for vocal characteristics with transcript content evaluation, the system captures both delivery and substance of responses. This comprehensive approach enables more accurate detection of potentially concerning responses by considering both {\emph what} participants say and {\emph how} they say it.

Our paper is organized as follows: We give a summary of the data we used, how each model was trained, the architecture of the system, and how we benchmark the system in \S \ref{sec:method}. In \S \ref{sec:results} we show the effectiveness of this new pipeline over the baseline approach. In \S \ref{sec:discussion}, we provide some natural directions and applications for this research. 

\section{Method}\label{sec:method}

\subsection{System Architecture}

\begin{figure}[!ht]
\begin{center}
\begin{tikzpicture}[xscale=1.5]
\draw[rounded corners=4pt, fill=yellow!5] (-3,.7) rectangle (-.5,5.2);
\node[rotate=90] at (-2.8,3) {Content Classifier};
\draw[rounded corners=4pt, fill=yellow!5] (2.8,.7) rectangle (.6,5.2);
\node[rotate=-90] at (2.6,3) {Prosodic Classifier};
\node[circle,draw=black, thick](or) at (0,4.5) {or};
\node[module, fill=blue!10](audio) at (0,0) {Audio};
\node[module, fill=red!10](trans) at (-1.5,1.5) {\begin{tabular}{c} Transcription\\Service \end{tabular}};
\node[module, fill=green!10](as) at (1.5,2) {\begin{tabular}{c}Audio\\Scorer\end{tabular}};
\node[module, fill=green!10](ts) at (-1.5,3) {\begin{tabular}{c}Text\\Scorer\end{tabular}};
\node[module, fill=yellow!10](tsc) at (-1.5,4.5) {\begin{tabular}{c}Cut-off\end{tabular}};
\node[module, fill=yellow!10](asc) at (1.5,4.5) {\begin{tabular}{c}Cut-off\end{tabular}};
\node[module, fill=blue!10](class) at (0,6.3) {Classification};
\draw[thick, rounded corners=5pt,->] (audio) -| (as);
\draw[thick, rounded corners=5pt,->] (audio) -| (trans);
\draw[thick, rounded corners=5pt,->] (as) -- (asc);
\draw[thick, rounded corners=5pt,->] (trans) -- (ts);
\draw[thick, rounded corners=5pt,->] (ts) -- (tsc);
\draw[thick, rounded corners=5pt,->] (tsc) -- (or);
\draw[thick, rounded corners=5pt,->] (asc) -- (or);
\draw[thick, rounded corners=5pt,->] (or) -- (class);
\end{tikzpicture}
\end{center}
\caption{The system is defined by two parallel processes; a content classification and a prosodic classification. \label{fig:architecture}}
\end{figure}
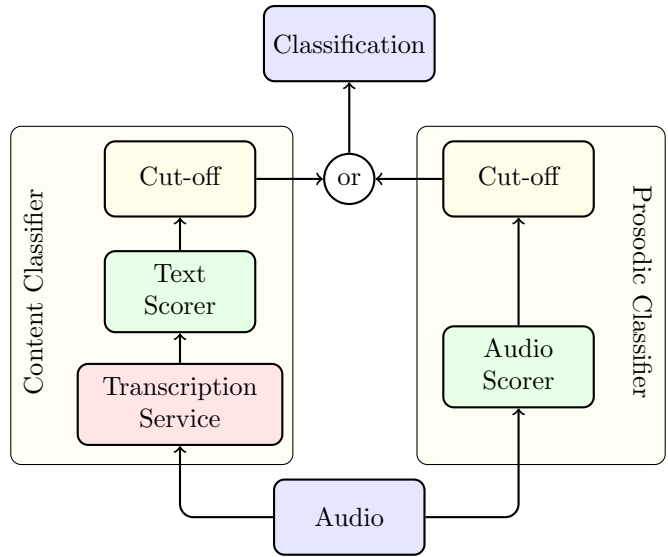

At a high level, the system contains three main components; a transcription service, a text scorer, and an audio scorer. The transcription service converts the audio to text which is used as input into the text-scorer, while the audio scorer is applied to the audio directly. The outputs of the text-scorer and audio scorer are real numbers. When we apply cut-offs to the outputs, we obtain two classifications. Our combined pipeline classifies the audio as an alert if either of our classifiers judges the audio to be an alert. This flow is presented in Figure \ref{fig:architecture}.

This means that we have two classifiers; one based on the transcription, which we call the content classifier, and one based directly on the audio, which we call the prosodic classifier. Together, they classify {\emph what} is being said, and {\emph how} they say it. 

\subsection{Data}

To define our task, we first reference the guidelines the hand-scoring team use to identify alerts. It is worth noting that many testing agencies have differing definitions of what constitutes an alert. The Smarter Balanced Consortium Hand-scoring rules identify ``Troubled Student Alerts" to include instances of suicide, criminal activity, alcohol or drug use, extreme depression, violence, rape, sexual, or physical abuse, self-harm or intent to harm others, or neglect. The hand-scoring team we used to identify alerts classifies alerts into five general categories; harm to self, harm to others, harm from others, severe depression, and a specific request for help. The clearest description of these categories can be found in the work of Burkhardt et al. \cite{burkhardt_rubric_2021}. 

As motivation for this study, many of these categories can be judged on the textual content of the speech alone. The elements of communication that are not covered by transcriptions are known as (vocal) prosody, which include elements like stress, tone, rhythm, and tempo. There are well-documented changes in vocal prosody that are linked with depression \cite{yang_detecting_2013}. Clearly, any comprehensive system for detecting alerts must take prosody into account. This motivates our use of a content classifier and a prosodic classifier.

Given we require a textual classifier, we first specify a corpus of textual responses. In \cite{ormerod_neural_2018} and \cite{ormerod_using_2023}, there is a corpus of appropriate text responses for classifying alarming student responses. As discussed in \cite{ormerod_using_2023}, there are approximately one alert for every 8,000 responses, which means that they are incredibly rare. In order to sample the number of alerts at a reasonable rate for the purposes of training a classifier, the set of alarming student responses is complimented by a set of supplementary responses vetted by a hand-scoring team as responses that would satisfy the criteria defined by the hand-scoring rules. Since these responses have no associated audio, they serve to train the content classifier. The details for this corpus are presented in Table \ref{traindat}.

\begin{table}[!ht]
\begin{center}
\begin{tabular}{l  l | r r r} \toprule
Category & & Alert & Normal  & Total \\ \toprule
Training & Student & 20,409 & 1,214,381  &  1,234,790 \\
 & Supplementary & 5,476 & 4,122 &  9,598  \\ \midrule 
 & Total & 25,885 & 1,218,503&  1,244,388\\\bottomrule
\end{tabular}
\end{center}
\caption{This data represents the text training data used in this study.\label{traindat}}
\end{table}

 The prosodic classifier is trained on audio data as described in Table \ref{traindat:audio}. Along with alarming student response, the responses used for the non-alert data were drawn from the typical types of items we expect to see alerts from.

\begin{table}[!ht]
\begin{center}
\begin{tabular}{l | r r r} \toprule
Category & Alert & Normal  & Total \\ \toprule
Training & 291 & 9,800  &  9,991 \\\midrule
Validation & 100 &  86,783 & 86,883  \\ \bottomrule

\end{tabular}
\end{center}
\caption{The audio data used to build the prosodic classifier and validate the system.\label{traindat:audio}}
\end{table}

One of the problems we have at this time is that the set of training examples for troubled students is an order of magnitude smaller than the number of training examples for text. We therefore use only a subset of the available non-alert data, with 200 sampled audio responses per each of the 49 prompts (9,800) for training. At validation, we use a large (86,783) corpus of responses to approximate the distribution of outputs from both the text scorer and the audio scorer. While this corpus is audio, we need to establish the distribution of the entire pipeline, not just the distribution of scores from text. 

\subsection{Models}

Transformer-based architectures (see \cite{vaswani_attention_2017}) have proven to be effective in multiple modalities including language \cite{devlin_bert_2018}, speech \cite{radford_robust_2022}, and vision \cite{dosovitskiy_image_2021}. They also prove to be effective in representation learning, allowing for learning to take place between different modalities \cite{bengio_representation_2013}. The architecture described above requires three models; a speech-to-text model, as part of an ASR pipeline, a text-classifier that classifies the text output of the speech-to-text model, and an audio-classifier that classifies the audio directly.   To be more specific, we have three transformer-based models we chose are versions of the Whisper models from \cite{radford_robust_2022} and a version of the ELECTRA model \cite{clark_electra_2020}. The transcriber model was not fine-tuned for the task. These choices, and basic descriptions of these models, are presented in Table \ref{tab:transformer_models}.

\begin{table}[]
    \centering
    \begin{tabular}{c | p{2cm} | c | p{8cm}} \toprule
     Function & Model &  Ref. & Description \\ \midrule
     Transcription &   Whisper Large (V3) & \cite{radford_robust_2022} & A 1.5 billion parameter model with 32 (transformer) layers. The inputs are Log-Mel Spectrogram transformed audio and the outputs are text.\\
     Audio Classifier &  Whisper Medium Classifier &  \cite{radford_robust_2022}  &  A 769 million parameter model with 24 (transformer) layers. The inputs are Log-Mel Spectrogram transformed audio and the outputs are real-valued elements of the interval $[0,1]$. \\
     Text Classifier & ELECTRA (small) Classifier & \cite{clark_electra_2020} & A 13 million parameter model with 12 (transformer) layers, adversarially trained as a discriminator (see \cite{clark_electra_2020}). The inputs are tokenized texts, and the outputs are real-valued elements of the interval [0,1]. \\ \bottomrule
    \end{tabular}
    \caption{The transformer-based models used in the classification pipeline.}
    \label{tab:transformer_models}
\end{table}

The text scorer was trained using the Adam classifier with weight decay \cite{loshchilov_decoupled_2019} with a learning rate of $5\times 10^{-6}$ applied to the cross-entropy loss function over 2 epochs over the entire dataset. Similarly, the audio scorer was trained with the same optimizer, loss function, and number of epochs with a learning rate of $5\times 10^{-6}$. As the loss function applies to the log-probabilities, the final score is the component associated with the probability of an alert when the log-probabilities are passed through a softmax function, meaning the final score has an interpretation as the probability of being an alert. 

\subsection{Benchmarking}

The system is designed to classify a particular percentage of all responses for review by a team of humans. There are mainly cost considerations to take into account when choosing that percentage. For this study, we report a range of percentage values that make sense between 0.3\% and 4\%, however, the we typically chose between 1\% and 2\%. 

Following the system architecture, the content classifier and prosodic classifier require cut-off values associated with those percentage. Since the classifiers are defined by providing cut-off values to the content and prosodic scorers, the performance of the classifiers can be determined independently. Suppose that $X$ is the set of responses, then we denote the application of the transcription followed by the text scorer function by $f_c : X \to [0,1]$. The application of the audio scorer is denoted by $f_p : X \to [0,1]$. The set of validation responses allows us to approximate the percentile function fairly accurately using linear interpolation, which provides us with approximations for the cut-off values, $c_c$ and $c_p$. That is to say, 
\[
P\left(f_c(x) > c_c | x \in X\right) = \frac{p}{100} \hspace{1cm} \textrm{and} \hspace{1cm} P\left(f_p(x) > c_p | x \in X\right) = \frac{p}{100}
\]
where $p$ is one of the values between $0.3$ and $4$ discussed above. 

In setting the cut-off values when considering the combination of the content and prosodic classifiers, our key assumption is that the percentage of responses flagged by each classifier, $\tilde{p}$, is the same. Given that $\tilde{c}_c$ and $\tilde{c}_p$ are the associated cut-off values for a given percentage $\tilde{p}$, for a desired percentage, $p$, our goal is to find a $\tilde{p}$ such that
\[
g(\tilde{p}) = P(f_c(x) > \tilde{c}_c || f_p(x) > \tilde{c}_p | x\in X) = \frac{p}{100}.
\]
We can derive $\tilde{p}$ implicitly using numerical root finding applied to the equation $g(\tilde{p})-\frac{p}{100} = 0$. We used the secant method to define the two appropriate cut-off values for each chosen percentage value with an initial estimate of $\tilde{p} = p/2$, which assumes a negligible intersection between the two classifiers. Once appropriate cut-off values are found, we can use the alerts in our test set to estimate the percentage of alerts flagged for review.

\section{Results}\label{sec:results}

As described above, for any given percentage $p$ of the population flagged for review, we can calculate the number and percentage of alerts that are flagged by each methods: the content classifier alone, the prosodic classifier alone, and both classifiers combined. The effectiveness of each of these methods using the number of alerts flagged under these assumptions. For a set of reasonable values between $0.3\%$ and $4\%$, we present these numbers in Table \ref{tab:efficacy}.

\begin{table}[!ht]
    \centering
    \begin{tabular}{c| r r | r r | r r } \toprule
       & \multicolumn{2}{c|}{Prosodic Classifier} & \multicolumn{2}{c| }{Content Classifier} & \multicolumn{2}{c}{Hybrid Classifier}  \\
      Percentage Routed   & N & \% & N & \% & N & \% \\ \midrule
    0.3&  57 & 57.0\% & 51 & 51.0\%  & 66 & 66.0\%  \\
    0.5&  62 & 62.0\% & 55 & 55.0\% & 75 & 75.0\% \\
    0.7&  65 & 65.0\% & 56 & 56.0\% & 77 & 77.0\% \\
    1 & 69 & 69.0\% & 60 & 60.0\% & 79 & 79.0\% \\
    2 & 78 & 78.0\% & 66 & 66.0\% & 85 & 85.0\% \\
    4 & 84 & 84.0\% & 76 & 76.0\% & 91 & 91.0\% \\ \bottomrule
    \end{tabular}
    \caption{The efficacy results for the audio (prosodic) classifier and text (content) classifier individually and combined in the hybrid system.}
    \label{tab:efficacy}
\end{table}

While the validation sample only contains 100 alerts, we clearly see that the combination of content and prosodic classifiers outperforms either of the two approaches alone. The results demonstrate that combining content and prosodic classifiers significantly improves the detection of concerning student responses compared to using either classifier alone. At the operationally relevant range of 1-2\% of responses flagged for review, the hybrid system identifies 79.0-85.0\% of alerts, compared to 60.0-66.0\% for the content classifier and 69.0-78.0\% for the prosodic classifier alone.

\section{Discussion}\label{sec:discussion}

The results' substantial improvement in detection rates could mean identifying several additional students in crisis. The performance gap between the content and prosodic classifiers (roughly 10 percentage points across most thresholds) likely reflects two factors. First, our content classifier benefited from a larger training dataset, including supplementary examples. Second, some alert categories, such as specific threats, may be more reliably detected through content than prosody alone. However, the prosodic classifier's ability to identify alerts not caught by content analysis (as evidenced by the hybrid system's superior performance) validates our hypothesis that vocal characteristics provide crucial complementary information.

Several limitations of this study suggest natural directions for future research. The relatively small number of audio training examples with alerts indicates that collecting additional audio data could substantially improve the prosodic classifier's performance. Furthermore, while our current architecture treats the two classifiers as independent, future work could explore more sophisticated ways of combining their outputs, potentially using the confidence scores from each classifier to weight their relative contributions.

The system could also be extended to classify alerts into the specific categories outlined in Table 1, which would help prioritize responses for human review. Additionally, while we focused on English-language responses, the framework could be adapted for other languages, though this would require careful consideration of how prosodic markers of distress may vary across cultures.

From an implementation perspective, the system's modular architecture allows for straightforward updates as improved models become available. For instance, the recent rapid advances in ASR and language models suggest that both classifiers could benefit from newer architectures or pre-trained models as they are released.

Finally, while this work focuses on automated scoring contexts, the approach could be valuable in other scenarios where early detection of concerning responses is critical, such as mental health hotlines or counseling services. However, such applications would require careful validation and likely modification of the alert criteria and classifiers for these specific contexts.

\bibliographystyle{plain}
\bibliography{lib}
\end{document}